\pdfoutput=1

\documentclass{article}

\usepackage{setup}

\begin{document}



\def\TITLE{Application of Facial Recognition using Convolutional Neural Networks for Entry Access Control}

\def\GROUP{Deep\_Learning\_Group 11}

\def\AUTHORS{
    Lars Ankile (larslank) \\
    Morgan Heggland (morganfh) \\
    Kjartan Krange (kjartkra) \\
}


\begin{titlepage}

\vbox{ }

\vbox{ }

\begin{center}
\includegraphics[width=0.40\textwidth]{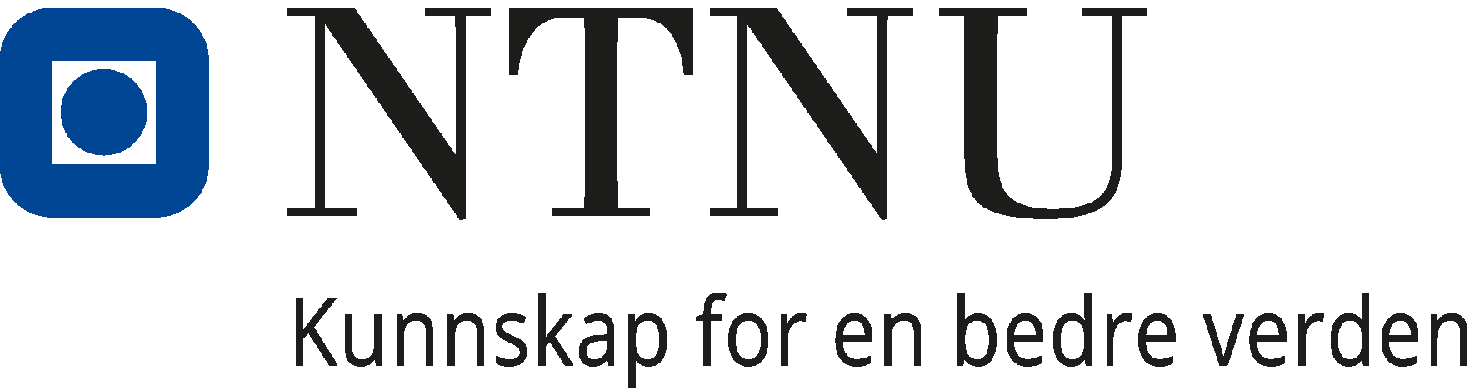}\\[1cm]
\textsc{\LARGE Department of Computer Science}\\[1.0cm]
\textsc{\Large TDT4173 - Final Project Paper}\\[0.5cm]

\def\checktitle{Paper Title}
\ifx\TITLE\checktitle
    \todo[inline]{
        Remember to fill in paper title, group name, and members in "title.tex"
    }
\fi

\vbox{ }
\HRule \\[0.4cm]
{ \huge \bfseries \TITLE}\\[0.4cm]
\HRule \\[1.5cm]
\large

\emph{Group:}\\
\GROUP

\emph{Authors:}\\
\AUTHORS

\vfill
{\large \today}
\end{center}
\end{titlepage}






\frontmatter

\begin{abstract}
    The purpose of this paper is to design a solution to the problem of facial recognition by use of convolutional neural networks, with the intention of applying the solution in a camera-based home-entry access control system. More specifically, the paper focuses on solving the supervised classification problem of taking images of people as input and classifying the person in the image as one of the authors or not. Two approaches are proposed: (1) building and training a neural network called WoodNet from scratch and (2) leveraging transfer learning by utilizing a network pre-trained on the ImageNet database and adapting it to this project's data and classes. In order to train the models to recognize the authors, a dataset containing more than 150 000 images has been created, balanced over the authors and others. Image extraction from videos and image augmentation techniques were instrumental for dataset creation. The results are two models classifying the individuals in the dataset with high accuracy, achieving over 99\% accuracy on held-out test data. The pre-trained model fitted significantly faster than WoodNet, and seems to generalize better. However, these results come with a few caveats. Because of the way the dataset was compiled, as well as the high accuracy, one has reason to believe the models over-fitted to the data to some degree. An added consequence of the data compilation method is that the test dataset may not be sufficiently different from the training data, limiting its ability to validate generalization of the models. However, utilizing the models in a web-cam based system, classifying faces in real-time, shows promising results and indicates that the models generalized fairly well for at least some of the classes (see the accompanying video). Next steps include collecting a much larger and more diverse dataset and researching other loss functions that penalize false positives more, in order to account for the shortcomings of the models.

\vspace{1cm}
\textbf{Additional resources:}
\begin{itemize}
    \item \href{https://github.com/Neuralwood-Net/woodnet}{\color{blue}GitHub repository} containing source code for the project
    \item {\href{https://youtu.be/EOOOq1k8TvE}{\color{blue}Accompanying video}} demonstrating and explaining the project
    \item Links to the datasets\footnote{All individuals photographed in the datasets have given their written approval to publish the data.} used in the project:
    \begin{itemize}
        \item \href{https://storage.googleapis.com/tdt4173-datasets/faces/images/raw_images.zip}{\color{blue}Raw images} (5.2 GB)
        \item \href{https://storage.googleapis.com/tdt4173-datasets/faces/balanced_sampled_224px_color_156240_images_70_15_15_split.zip}{\color{blue} Final dataset with images cropped around center} (3.3 GB)
        \item \href{https://storage.googleapis.com/tdt4173-datasets/faces/balanced_sampled_cropped_224px_color_70_15_15_split.tar.gz}{\color{blue} final dataset with images cropped to faces} (3.1 GB)
    \end{itemize}
\end{itemize}
\end{abstract}
\clearpage

\tableofcontents

\listoffigures
\listoftables

\mainmatter

\section{Introduction}
\label{sec:introduction}
This paper addresses the supervised classification problem of \emph{facial recognition} using  \emph{convolutional neural networks} (CNNs). That is, given an image, classify the image into one of a finite set of classes, each representing a person. This problem is addressed with the intention of applying it to entry access control, an approach that could provide fast, easy, and secure, key-less access to e.g. an apartment. This is an interesting problem for a number of reasons. For one, the prevalence of biometric recognition is growing, highlighting the relevance of this paper. For instance, facial recognition systems are now commonplace in virtually all smartphones. Furthermore, it is said that there is nothing new under the sun – except the extensive dataset used in this project. Applying CNNs for image classification is not exactly new, however, this project's novelty stems from a meticulously constructed dataset, a network made from scratch, and the comparison with a pre-trained network. It is thus a novel application of facial recognition which, if successful, may yield very tangible and meaningful results (to the authors at least). This was a critical factor for the authors in choosing this project. 

To classify faces, a model that accepts images and outputs a prediction is created. The data used for this task is color images of people and other motifs. The model's goal is to classify the images into one of the four classes: $\{Kjartan, Lars, Morgan, Other\}$, where the three former classes represent the faces of the authors of the paper. Such a model may function as the discriminator in an automated face-based admission control system, where three people have approved access while the ``Other'' category represent trespassers - i.e. everyone else. One application of such a system could be controlling access of the door to an apartment the three authors share.

The data used for training has been gathered and processed by the authors. Collection starts by filming faces in different settings with varied conditions (e.g lighting). The videos are converted into separate images, and each image is processed and augmented to increase the amount of data. This way, the model may avoid discriminating classes based on spurious correlations with faces, like background, clothing, lighting etc. To gather data for each class, both the authors and other people are filmed in the differing settings. The steps taken to preprocess the data is a substantial part of the workload, and is described in further detail in \autoref{sec:data}.

In order to classify faces, a \emph{deep learning} model, specifically a CNN called \emph{WoodNet} is designed by the authors. CNNs are appropriate for facial recognition as they are well suited for analyzing grid-like data such as images through their distinctive convolution operation\footnote{The convolution operation is elaborated upon in detail in the method paper previously written by the authors.}. This enables CNNs to exploit several invariant properties of the data such as \emph{translational invariance} and \emph{spatial locality}, allowing for identification of patterns in images, such as features of a face. \emph{Parameter sharing} makes CNNs small compared to a comparable dense neural network. In order to train the network, \emph{Stochastic Gradient Descent} (SGD) with \emph{backpropagation} is utilized, a well-documented method for optimization of the network. See \autoref{sec:methods} for details.

In addition to implementing a CNN, the machine learning method \emph{transfer learning} is also applied. Numerous publicly available neural networks, trained with large amount of images (e.g ImageNet with 10 million images over 10 000 categories) exist. Transfer learning is utilized by modifying the final layer of such a large and well-trained net. This allows for rapid retraining while leveraging the accurate feature extraction of the trained model. In this paper, a condensed version of AlexNet \citep{NIPS2012_c399862d}, called SqueezeNet \citep{iandola2016squeezenet} is chosen. See \autoref{sec:related-work} and \autoref{sec:methods} for further details.

Lastly, the authors would like to emphasize that working on this project has meant iteratively experimenting on countless different models, data processing approaches and hours upon hours of training without seeing many results. Since this paper is part of the NTNU course ``TDT4173 - Machine Learning'', the different attempts performed are supplied in the \href{https://github.com/Neuralwood-Net/woodnet/tree/main/notebooks/archive}{\color{blue}\texttt{notebooks/archive}} folder in this project's GitHub repository to document the learning process.

\section{Related Work}
\label{sec:related-work}
\begin{figure}
    \centering
    \includegraphics[width=0.8\textwidth]{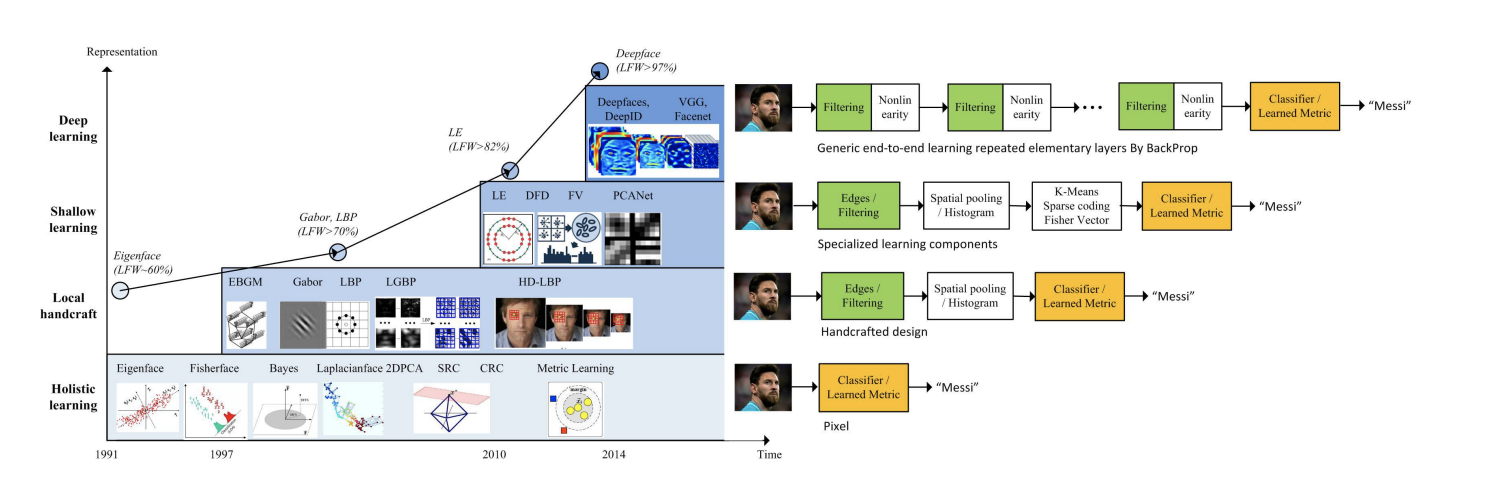}
    \caption[Historical overview of facial recognition approaches.]{Historical overview of facial recognition approaches.}
    \source{\cite{DBLP:journals/corr/abs-1804-06655}}
    \label{fig:fr-overview}
\end{figure}

This section contextualizes this paper's project within the field of deep learning, with emphasis on CNNs and facial recognition. Presentations of related work is divided into two groups: first, outline of important research for deep learning generally, and second, an overview of crucial contributions to the problem of facial recognition. As \autoref{fig:fr-overview} shows, over the last three decades, the state-of-the-art facial recognition algorithms' accuracy have improved in correlation with an increasing prevalence of deep learning approaches and CNNs.

\emph{Convolutional neural networks} \\
The breakthrough application of CNNs on the MNIST dataset in 1998 \citep{LeCun1998GradientbasedLA} marked the beginning of a revolution in using neural networks for complex machine learning, in particular image recognition, classification, detection, and segmentation. Ever since \emph{AlexNet} \citep{NIPS2012_c399862d} won the \emph{ImageNet Large Scale Visual Recognition Challenge} (ILSVRC) competition in 2012, CNNs and deep learning have proven to be the most proficient methods for image classification. Another method used in this paper, which is important for the evolution of the field of deep learning, is SGD \citep{MIT-6-036}, which computes the gradient of the parameters using only a random sample of training examples. \emph{Adam} \citep{Adam} is a slightly more complex and adaptive variant that often leads to more efficient training, and is discussed in \autoref{sec:methods}.

\emph{Facial recognition} \\ AlexNet also contributed to the field of facial recognition specifically. It was used as an underlying architecture for many of the first deep learning facial recognition algorithms implemented in the years around 2013 \citep{DBLP:journals/corr/abs-1804-06655}. In this paper, a smaller\footnote{In terms of number of weights.} version of AlexNet, \emph{SqueezeNet} \citep{iandola2016squeezenet} was used as the base for the transfer learning approach and proved to achieve the higher accuracy faster (see \autoref{sec:Transfer-Learning}).

One highly relevant facial recognition algorithm in relation to this paper, is the \emph{FaceNet} model, developed by Google researchers \citep{DBLP:journals/corr/SchroffKP15}. It still is one of the most accurate facial recognition algorithms\footnote{\href{http://vis-www.cs.umass.edu/lfw/results.html}{\color{blue}http://vis-www.cs.umass.edu/lfw/results.html}}. FaceNet uses a large private database of faces, and complex loss function to train a GoogLeNet \citep{DBLP:journals/corr/SzegedyLJSRAEVR14} to extract face embeddings. It achieves a staggering 99.63\% accuracy on the \emph{Labeled Faces in the Wild} (LFW) database which consists of 13,000 labeled pictures \citep{LFWTech}. The paper also shows that the sizes of the datasets continue to matter for large sample sizes. Using tens of millions of images resulted in a clear boost of accuracy as compared to ``only'' millions of images. Over this interval the relative reduction in error is 60\%. The data collected for this project consists of approximately 150 000 images (see \autoref{sec:data}). Experiences from FaceNet might indicate that this project could have benefited from gathering a significantly larger dataset.
\section{Data}
\label{sec:data}
To classify the authors' faces, a dataset with 156 241 color images of size 224 by 224 pixels have been produced. There are $156\,240 / 4 = 39\,060$ images for each of the four classes $\{Lars, Morgan, \\ Kjartan, Other\}$.  The images are split into a training, validation, and test set in parts of 70\%, 15\%, and 15\% or 109 368, 23 436, and 23 436 images, respectively. This split is chosen because it provides a reasonable trade-off between sufficient test and validation data and having enough data to train on. In testing the models developed during the project, two versions of the dataset was produced for comparison: one that uses face detection to crop out faces, and one that crops the image around the center, making it square. 

\begin{figure}
    \centering
    \includegraphics[width=0.7\textwidth]{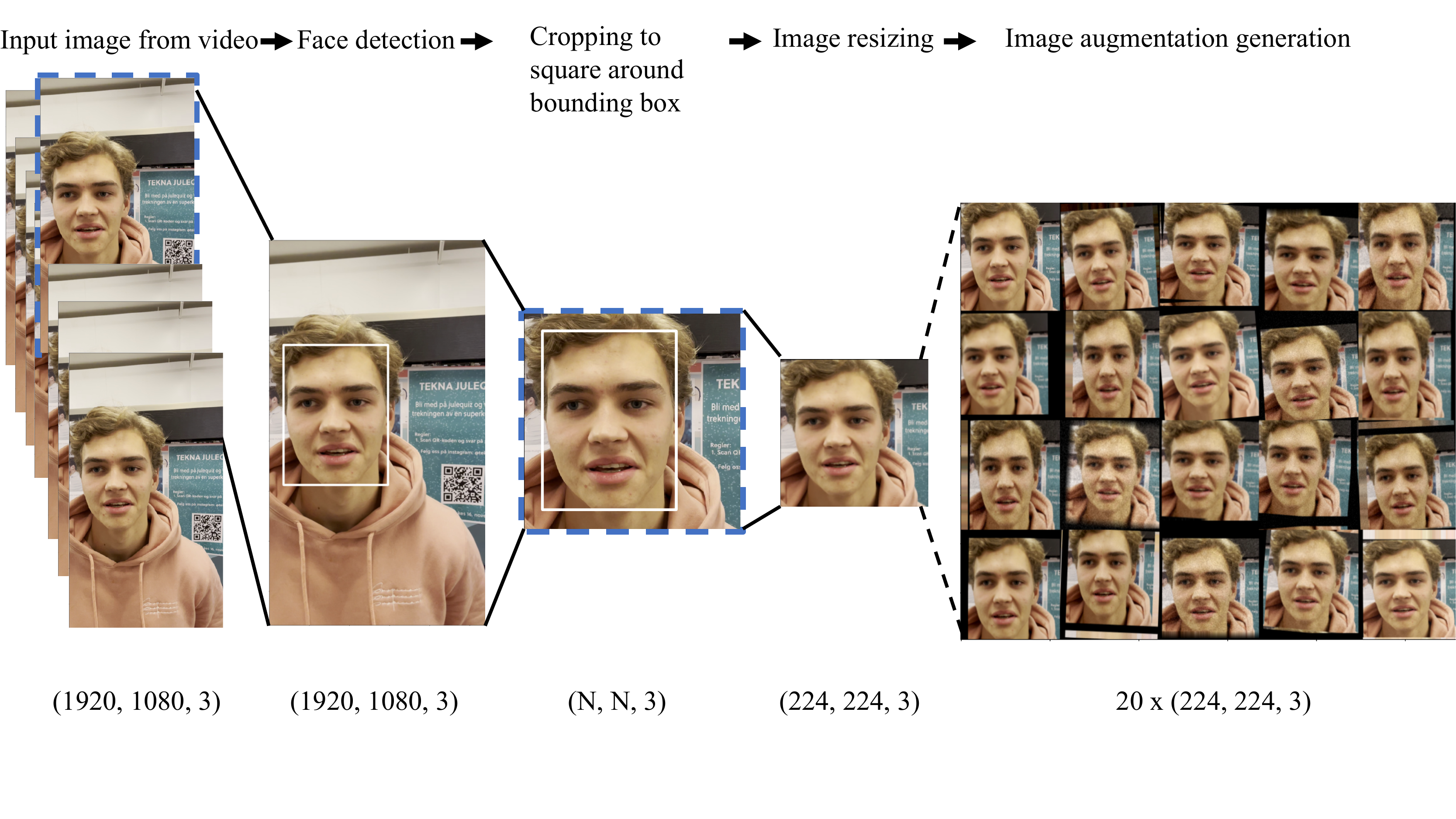}
    \caption[The image processing pipeline used to generate training data.]{Pipeline for generating training data. An alternative pipeline without face detection was also used, where images were cropped to square around the center (see \autoref{subsec:compression}). The numbers below the images give the dimensions at each step. N is any size and depends on the size of the face in any given image.}
    \label{fig:pipeline}
\end{figure}

As the raw data is high resolution videos, a number of processing steps are required to create a dataset ready to fit a model with. This section outlines the collection and transformations of the data used. See \autoref{fig:pipeline} for a visual overview. The raw images and the finished datasets can be found under ``additional resources'' below the abstract.

\subsection{Raw Data Collection}
Data collection was performed by recording videos of each of the authors' and their friends' faces repeatedly, with a variety of backgrounds and light settings. Video recordings were used to capture high volumes of images efficiently. The videos were converted to separate images, each with one person in focus. While recording, the captured face was centered, while keeping nothing else in the frame constant. This was done by capturing videos with different backgrounds, surrounding colors, lighting, clothes etc. This way, data that induces learning of facial features specifically should be generated. The variation should prevent the neural networks from learning to classify individuals by, for instance, a dazzling sweater. More formally, given two images of the same class, the ambition is for their intersection to exclusively be some structure the computer can recognize as a single individual's face.

In addition to images of friends of the authors, the \emph{Other} class also contains pictures without people in them in an effort to make the networks learn to predict this class when there is no person in the frame. In an entry access control scenario, this ensures that the system does not wrongfully activate and allow access when there is no one in front of the camera. 

As the data collection resulted in a varied number of images per class, the dataset was balanced before further processing was conducted. Let $n$ denote the number of images in the class containing the fewest images. In order to balance the dataset, a random sample of $n$ images were extracted for each of the three other classes. This resulted in 7812 images per class before augmentation. Balancing prevents bias towards heavier classes and prevents reduced predictive performance regarding the underrepresented classes.
 
\subsection{Compression}
\label{subsec:compression}
For efficient storage, and more importantly, rapid training, the images were compressed to a smaller size. This is a trade-off between information and performance, as one has to balance information with data size and network size. An image size of 224x224 was chosen after iterative experimentation, where several different sizes were tested. For images of size 64x64, there was clearly a substantial loss of information, and even the human eye struggled with classification. At 128x128, the human eye could easily distinguish between classes, but training the model to do so proved challenging as low accuracies were achieved. At 224x224 pixels the images seem detailed enough for efficient feature extraction. At the other extreme, images of size 1080x1080 (length of short end of original images) would increase use of space and complexity by a factor of $\frac{1080^2}{224^2}\approx 23$. Lastly, since ImageNet images are 224x224, that further reinforces the assumption that this size is a good trade-off between information and size. The preprocessing of the data was done in three parts. These are detailed below, accompanied by reflections around and justifications behind each processing step. The code used to perform this processing can be found in \href{https://github.com/Neuralwood-Net/woodnet/blob/main/notebooks/data-extraction.ipynb}{\color{blue}\texttt{notebooks/data-extraction.ipynb}}.

\begin{figure}
    \centering
    \includegraphics[width=.4\textwidth]{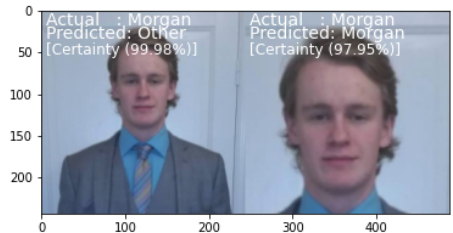}
    \caption[Demonstration of separating facial detection and recognition.]{Demonstration of separating facial detection and recognition. WoodNet does not recognize Morgan to the left, and is 99.98\% certain it is the \emph{Other} class (i.e. less than 0.02\% certain that it \textit{is} Morgan). However, on the right, WoodNet is 97.95\% certain it is him.}
    \label{fig:morgan-zoom}
\end{figure}

\begin{enumerate}
    \item \textit{Face detection.} To remove background and focus on the faces, faces were detected using a third-party pre-trained CNN, an implementation of the \emph{Multi-Task Cascaded CNN} (MTCNN) as described in \cite{mtcnn}. See \autoref{fig:pipeline} for an example of face detection applied to the collected data. The intent is to crop the image around the detected face (see step 2). This should improve the model's learning rate as it allows the neural network to train only on what is important: faces. The problem of detecting a face is a vastly different task to that of recognizing (classifying) one, and therefore using the pre-trained network to bound the face in each image seems suitable. As \cite{DBLP:journals/corr/abs-1804-06655} shows, separating the task of extracting a face from an image from the task of matching a face with a name is a common practice for even the most proficient facial recognition models. Furthermore, feeding WoodNet two considerably similar photos yielded two different predictions, where only one was correct, see \autoref{fig:morgan-zoom}. This illustrates how different the task of face detection and recognition is, and that cropping faces seems to be conducive for learning.
    
    \item \textit{Crop image to square.} After detecting a face in the image, the image is cropped around the bounding box yielded from the face detection CNN to produce a square picture of an individual's face. While obtaining the benefits of removing background, cropping the image helps reduce file sizes. For the dataset without face detection, this is the first step of the compression processing, and cropping is done by simply removing pixels from the edges of the image to make it square.
    
    \item \textit{Resizing.} As a final step, all images were resized to the size of 224 by 224 RGB pixels. In most cases, this resize is a downsize, as the input image will only be smaller in the case where the face detection algorithm detected a face far away. The resizing is primarily done to save resources, reduce training time and interface with SqueezeNet, although most likely at the cost of losing some valuable information in the data.
\end{enumerate}

\subsection{Augmentation}
 The last step in the preprocessing pipeline is image augmentation. After collecting the data and compressing it, there are 7 812 224x224 RGB images. As the images stem from videos, many of the individual images are similar. One can imagine a neural network overfitting to images that appear quite similar. Furthermore, it would be problematic if the dataset consisted of a small number of original images. To mitigate both of these issues, each image is randomly augmented in order to grow both the size and diversity of the dataset. Each original image is augmented 19 times, yielding 20 images per original image. Every augmentation include the same types of manipulations, however they are applied in random order and with random parameters within their respective ranges. The following transformations were used: Rotation of $\pm 5^{\circ}$, scaling the image by $-5\%$ to $+10\% $ by cropping and padding, additive Gaussian noise using a kernel with $0<\sigma\leq1$, changing the brightness from $-10$ up to $+10$ of the original value, and finally translating the image in $x$ and $y$ direction by $-10\%$ to $+10\%$ per axis. The transformations and their parameters were chosen on the basis that they sufficiently altered the picture without making it unrecognizable for the human eye.
 
 The right side of \autoref{fig:pipeline} shows an example image that has gone trough these transformations, creating 19 augmentations. By generating 19 new augmentations and including the original image, the dataset size is increased to $7\,812\cdot 20 = 156\,240$ images. The choice of 19 augmentations was made as a result of a qualitative evaluation of the trade-off between a larger dataset and more similar images. Naturally, by producing more augmentations per image the size of the data could be considerably increased, but at the risk of new images being a regurgitation of old ones.
\section{Methods}
\label{sec:methods}

The high level approach used in this paper is a deep convolutional neural network, where loss is measured by \emph{cross-entropy loss}, optimized with the stochastic gradient descent variant Adam, and trained on a dataset generated by the authors (see \autoref{sec:data}). Transfer learning by adapting a fully-trained network to this specific problem is also tested. This section elaborates on these methods and their rationale. In \href{https://github.com/Neuralwood-Net/woodnet/blob/main/notebooks/training-and-plotting.ipynb}{\color{blue}\texttt{notebooks/training-and-plotting.ipynb}} one can see and run the code used for defining the networks, training and plotting.

\subsection{Convolutional Neural Network Architecture}
As discussed in \autoref{sec:related-work}, CNNs have for a decade been dominant in most machine learning applications involving data with some aspect of temporal or spatial locality. As the problem for this paper is an image classification problem, the CNN is a natural choice.

To ease the implementation effort, the project relies on the deep learning framework \emph{PyTorch} (version 1.7.0+cu101)\footnote{\href{https://pytorch.org/}{\color{blue}https://pytorch.org/}}. This framework is chosen for several reasons. First, from the authors' research, it appears to be the most beginner friendly and pythonic, i.e. it feels most like programming pure python. Second, it is used by companies like Tesla to power the neural nets in their self-driving cars\footnote{\href{https://www.youtube.com/watch?v=oBklltKXtDE}{\color{blue}PyTorch at Tesla - Andrej Karpathy, Tesla}}, and should therefore be sufficient for this undertaking as well.

\begin{figure}
    \centering
    \includegraphics[width=0.7\textwidth]{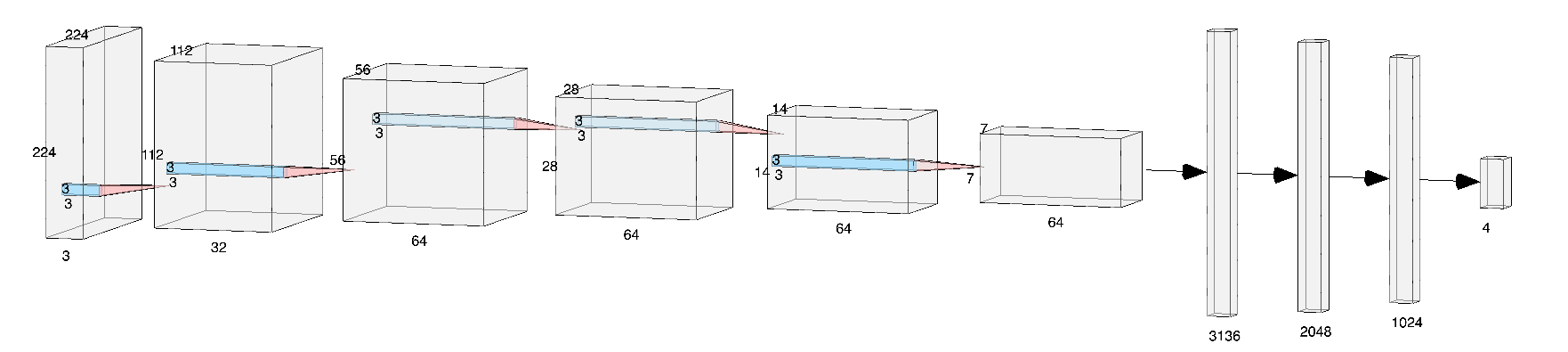}
    \caption[Visualization of the WoodNet architecture.]{Visualization of the architecture of WoodNet. The first 5 boxes from the left represent convolution units with given heights, widths, number of channels and filter sizes. The final four boxes to the right represent fully connected layers resembling arrays.}
    \label{fig:architecture}
\end{figure}

Through iterative experimentation, a network architecture prioritizing understanding, simplicity, and learning was designed. This was prioritized over creating the absolute best-performing network achievable. Certainly, when taking \autoref{sec:related-work} and for instance FaceNet into consideration, the fact that the neural network was trained using thousands of hours on Google CPU clusters \citep{DBLP:journals/corr/SchroffKP15}, with the company's private database of faces in the hundreds of millions, makes a goal of matching their accuracy seem unreasonable for this project. This resulted in the implementation of a fairly simple architecture. Inspiration for possible architectures has been drawn from distinguished architectures such as FaceNet \citep{DBLP:journals/corr/SchroffKP15}, VGG \citep{simonyan2015deep}, AlexNet \citep{NIPS2012_c399862d}, SqueezeNet \citep{iandola2016squeezenet}, etc. Simple architectures that solve the MNIST challenge were also researched. The end result is an architecture in between the two extremes. In short, the development started with a very simple network, see e.g. the code for ``BadNet''\footnote{Tongue in cheek name for a network that performs a lot worse than the final architecture of WoodNet.} in \autoref{code:badnet} in \autoref{app:badnet} of the appendix. This architecture did not prove sufficient for efficient learning. Iteratively, the net was extended with more layers while monitoring the performance on validation data until a sufficiently deep network was found in WoodNet. Kernel size and number of dropout layers (see \autoref{sec:activation-function}) were also varied. As an example, WoodNet with dropout after every layer was much less apt for learning than the version with dropout only after the second-to-last layer. The PyTorch code defining the final network, WoodNet, is given in \autoref{code:woodnet} in \autoref{app:woodnet} of the appendix, and a visualization of the architecture can be seen in \autoref{fig:architecture}.

WoodNet consists of a feature extraction phase and a classification phase. The feature extraction phase is centered around triplets of the two-dimensional convolution layer, \emph{max pooling} \citep{Riesenhuber1999HierarchicalMO}, and a \emph{Rectified Linear Unit} (ReLU) non-linear activation \citep{ramachandran2017searching}. There are four sets of these triplets. The convolution layer is the distinctive element of a CNN \citep{Goodfellow-et-al-2016}. The max pooling cuts the height and width of the image in half for every convolution unit, but will, in theory, output more and more refined features. In short, an image of 224 by 224 pixels by 3 channels is fed in, and iteratively the model attempts to extract larger and larger features and ultimately outputs them in the form of a $7\times 7\times 64$ tensor. The max pool layer is chosen over e.g. average pooling because by allowing only the maximum value through the filter, it will emphasize the strongest signal which can help in propagating features through the network.

The output tensor from the feature extraction phase is then passed to the classification layer. This consists of three fully connected layers. The first layer takes in a flattened tensor of length $7\cdot 7\cdot 64 = 3136$, representing features if working as intended. The input layer consists of 2048 neurons, and is connected to the hidden classification layer that has 1024 neurons, before it is passed to the final layer that has four neurons, i.e. one for each class in the dataset. Every fully connected layer except the last is followed by a ReLU non-linearity and a dropout layer \citep{labach2019survey}.

\subsection{Choice of Activation Function and Regularization}
\label{sec:activation-function}
To ensure that the neural network is actually a function approximator that can approximate non-linear functions, the network requires non-linear activation functions \citep{MIT-6-036}, i.e. functions that take the output of one layer and performs a non-linear transformation to it. The ReLU is such a function and is defined as $\textnormal{ReLU}(x) = \max(0, x)$ \citep{ramachandran2017searching}. This activation function is chosen as it is simple and easy to understand and its derivative is trivially calculated, which speeds up training. Furthermore, it has proven effective in practice and is utilized by all the successful architectures mentioned previously. The final layer does not have an activation function. The \emph{Softmax} function could have been used, but was not necessary due to the choice of loss function (see \autoref{subsec:loss-func-optimization}). The Softmax function is however used to normalize the logits the network outputs to be able to interpret the model outputs as probabilities, when appropriate.

In an effort to make the network generalize and not be overly fitted to the training data, the network incorporates \emph{dropout} layers. Dropout prevents overfitting by preventing co-adaption of the network parameters on the data \citep{hinton2012improving}. This is achieved by randomly turning off connections between neurons during training. This way, it is theorized, that the remaining neurons are forced to generalize to the data, and over time the ``signal'' will emerge out of the noise. In the network, dropout is implemented in between the two last fully connected layers. Too much dropout can hinder the network from learning anything at all, as the authors experienced during experimentation.

\subsection{Loss Function and Optimization Criterion}
\label{subsec:loss-func-optimization}
The basic, and most widely used approach to training neural networks is stochastic gradient descent (SGD) \citep{MIT-6-036}. This involves calculating gradients of weights in the network, to figure out how they can be nudged towards an optimum. This requires a way to quantify how well or poorly the network performs at the classification task, i.e. a \emph{loss function}. The loss function solves this by providing a quantitative measure of how bad a certain prediction is. When this function is differentiated with respect to the parameters in the model, with the help of the chain rule for derivatives, one can use the resulting gradient to decide how the parameters in the network can be tuned to most efficiently decrease the loss, and thus increase the accuracy. This paper uses the \emph{Cross Entropy Loss} or \emph{Negative Log Likelihood Loss}. Cross entropy loss is useful for classification problems, generalizes over an arbitrary number of classes, and is easily differentiated \citep{Goodfellow-et-al-2016}. As this task is a classification task, cross entropy loss suits the problem well. Given $N$ input samples, $M$ classes, binary indicator $y_{i,c}$ stating whether observation $i$ is classified as class $c$ and $p_{i,c}$, the predicted probability of observation $i$ being of class $c$, the multi-class cross entropy loss function can be written as 

\begin{align*}
    H_p(q)=-\frac{1}{N}\sum_{i=1}^{N}\sum_{c=1}^{M} y_{i,c} log(p_{i,c}). \quad \text{\citep{MIT-6-036}}
\end{align*}

As mentioned, the network will be trained using SGD. However, there are variations of the classic algorithm that provide some advantages. In normal SGD, the step size used to adjust the weights are shared between all weights, and is normally decayed at some fixed rate. In this project, Adam is used instead, which utilizes adaptive estimates of the \emph{moment}, or change in loss to change in parameters, to more accurately set appropriate learning rates for efficient learning \citep{kingma2017adam}. Furthermore, the authors argue that Adam is computationally efficient, requires little memory, and is well suited for problems requiring a large number of parameters or data. Lastly, they argue that it outperforms most other stochastic optimization for neural networks available. It was even designated as the default algorithm to use by Andrej Karpathy at Stanford University in the course \textit{CS231n: Convolutional Neural Networks for Visual Recognition\footnote{\href{http://cs231n.stanford.edu/}{\color{blue}http://cs231n.stanford.edu/}}}. It was thus a natural choice for this project.

\subsection{Training Loop and Data}
The model trains on the dataset the authors created, as described in \autoref{sec:data}. Prior to the model ever seeing data, it is divided into three distinct set of images: training, validation, and test. The model is then trained on the training data and validated on the validation data after each epoch of training. An epoch is one full iteration through the training data. That means that the model will see the training data multiple times and will therefore be prone to over-fit it, and not generalize to the embedded features in them. That is why the model is tested on the validation set after every epoch. The accuracy on both the training data and the validation data is expected to increase in tandem for a number of epochs, while the model is learning. However, once the model starts overfitting, the accuracy on the training data can be seen to increase while the accuracy on the validation data decreases as a consequence of the model ``memorizing'' the images in the training data, and not learning the generalized features in the images. That way, one can keep track of the best model as measured by accuracy on the validation data, and save those parameters.

The held-out test data is meant as a last validation of model generalization. It will only be used once the model and hyperparameters are decided upon. This is a measure to prevent overfitting to the validation data, which might happen if one experiments with a large array of different networks, hyperparameters, and configurations.

\subsection{Transfer Learning for Better Performance with Less Data}
\label{sec:Transfer-Learning}
 Despite efforts in generating a large dataset specific for the task as discussed in \autoref{sec:data}, the dataset is in all likelihood too small and not diverse enough to enable a model to fully tease out specific features that are important for face recognition when trained from scratch. Therefore, the method of transfer learning is applied and contrasted with the results from learning from nothing. The network used for transfer learning is SqueezeNet \citep{iandola2016squeezenet}. This network is chosen because it shows similar performance to the well-known AlexNet on ImageNet, while having only one 50th the number of parameters and taking up 510 times less space. This suits this use-case because the network can be downloaded and fine-tuned quickly. The model is fully trained\footnote{\emph{``fully trained''} means that the model achieves state of the art performance on the given classification task, i.e. winning accuracy for ILSVRC 2012 of 57.5\% in this specific case.} on ImageNet with 1000 classes. To adapt it to the problem in this paper, the final classification layer will be replaced to predict for four classes instead. Then, the model is trained on the novel data. Fine-tuning the parameters in the final layer only will be experimented with. In theory, a model fully trained on ImageNet has learned how to effectively extract features from images, and will be able to effectively learn how to map those features to novel classes \citep{inductive-transfer}.

\subsection{Experiments and Evaluation}
The most important evaluation metric to be used to measure the performance of the models is the accuracy, i.e. percentage of correctly classified images. Furthermore, the loss value is recorded, as this is an indication of how ``bad'' the models are performing. In addition, precision and recall are interesting metrics given the context of access control. Considering a prediction of one of $\{Kjartan, Lars, Morgan\}$ as a positive prediction, and predicting $\{Other\}$ as a negative, precision and recall may be calculated. False positives represent allowing unauthorized access and the consequences of granting access for the wrong person are larger than not granting access for the right person. Thus, a high precision is desirable. Furthermore, a confusion matrix will be used to expand on this analysis, examining whether some classes are misclassified more often than others, and whether some classes are more often predicted as the result of a misclassification.

Another interesting experiment is to test the models with completely novel data, e.g. separate images of the authors from entirely different settings. An example of this is running live inference using an image stream from a webcam as input. Performance on these images will be a qualitative indicator of the models' ability to generalize. The dataset the networks are trained on is in all likelihood not sufficiently large or varied enough to enable the models to fully learn the features that define the faces of the authors. Therefore, it is to be expected that the accuracy on completely novel images is lower than that from the training, validation, and test set.
\section{Results} 
\label{sec:results}
This section presents the main results of training WoodNet and SqueezeNet using the two different datasets (images cropped to faces and cropped around center). Discussion and reflections regarding the results are also included. Experiments run to settle on the final dataset is discussed in \autoref{sec:data} and experiments deriving the final CNN architecture of WoodNet is discussed in \autoref{sec:methods}.

\subsection{Presentation of Results}
\label{sec:presentation-of-results}
A summary of the key statistical measures of the two networks for each test dataset is presented in \autoref{tab:key-metrics}. The test dataset was never tested on the networks before the final models were settled on for the higher validity. Overall, WoodNet achieves a marginally lower loss and higher accuracy than SqueezeNet. Notably, the face-cropped dataset achieves a higher loss and lower recall for both models, but in turn achieves a higher accuracy and precision.

\autoref{fig:training-plot} and \autoref{tab:training-table} in the appendix shows the accuracy and loss from training WoodNet and SqueezeNet on the face-cropped dataset. As a full epoch of training is completed before a full epoch of validation, validation always start at a better level than training. An observation to highlight is that both networks seem to learn fairly quickly, dropping loss to zero and accuracy close to 100\% after two epochs. Furthermore, SqueezeNet seems to learn significantly faster than WoodNet in terms of number of images seen. A final observation is that none of the graphs exert any signs of overfitting, which is normally visible through diverging graphs for training and validation.

\begin{figure}
    \centering
    \includegraphics[width=0.9\textwidth]{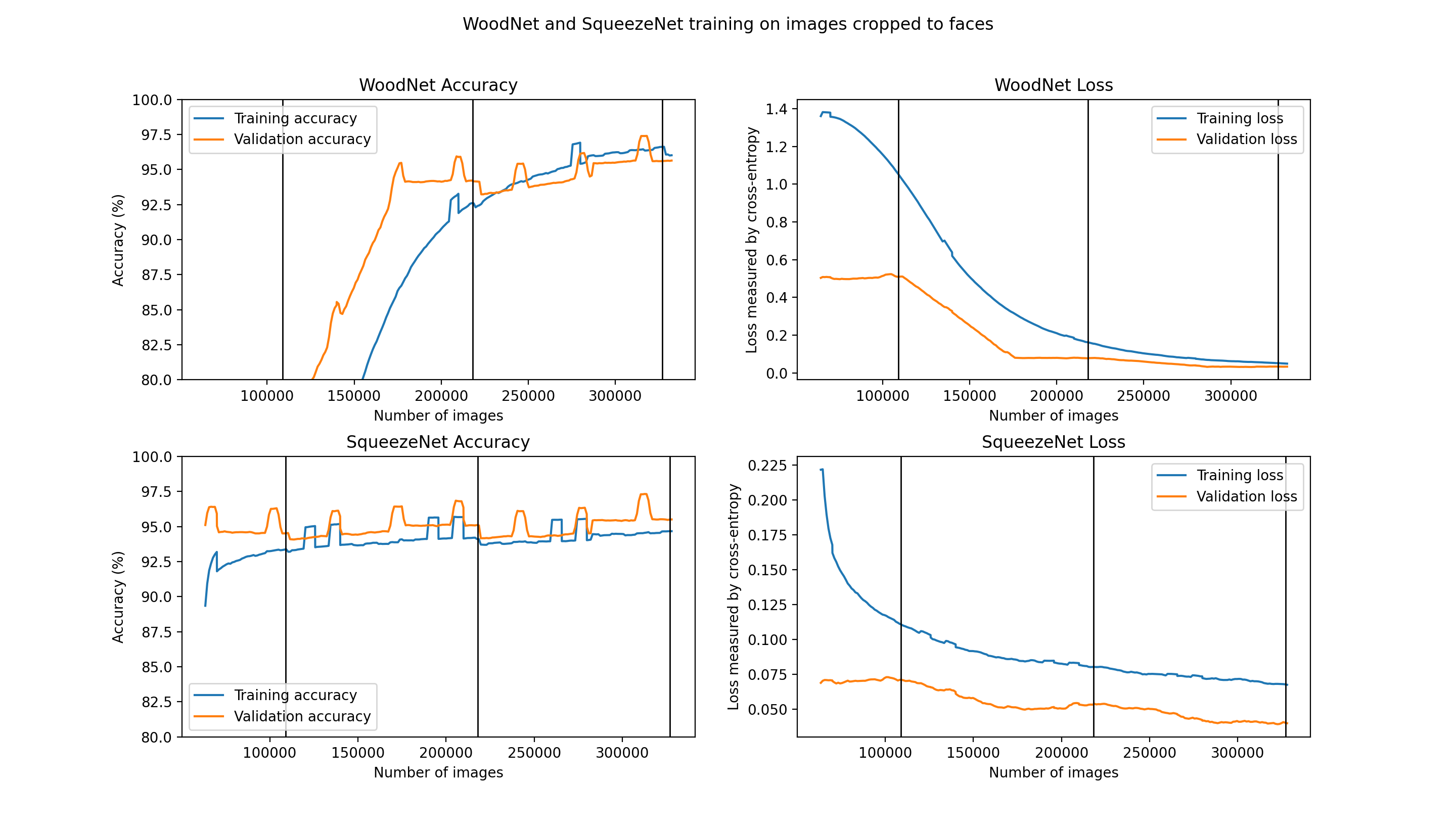}
    \caption[Training of WoodNet and SqueezeNet.]{Training of WoodNet and SqueezeNet. Each row shows the training of a given network. Each column shows the accuracy and loss respectively, both during training and validation, over the course of the first three epochs of training. The black, vertical lines marks the change of epoch. The $x$-axis is the cumulative number of images trained on.}
    \label{fig:training-plot}
\end{figure}

\begin{table}
\small
\centering
\begin{tabular}{l|ll|ll}
 & \multicolumn{2}{c|}{\textbf{WoodNet}} & \multicolumn{2}{c}{\textbf{SqueezeNet}} \\
\textbf{Metric} & \multicolumn{1}{c}{\textbf{Generic crop}} & \multicolumn{1}{c|}{\textbf{Face-crop}} & \multicolumn{1}{c}{\textbf{Generic crop}} & \multicolumn{1}{c}{\textbf{Face-crop}} \\ \hline
Loss & 0.00895 & 0.00985 & 0.02685 & 0.03501 \\
Accuracy & 0.99710 & 0.99780 & 0.99200 & 0.99020 \\
Precision & 0.99903 & 0.99960 & 0.99624 & 0.99685 \\
Recall & 0.99875 & 0.99852 & 0.99749 & 0.99305
\end{tabular}
\caption[Summary of important metrics for the two networks.]{Summary of important metrics for the two networks and the two sets of training data, calculated from the held-out test set that no network was tested on prior to settling on final models.}
\label{tab:key-metrics}
\end{table}

\begin{figure}
\centering
\includegraphics[width=0.7\textwidth]{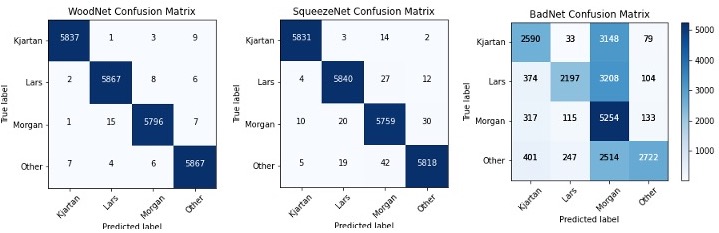}
\caption[The confusion matrices for each of the tested models.]{The confusion matrices for WoodNet, SqueezeNet and BadNet when tested on the generically cropped test dataset. The $x$-axis represents what the model predicts, and the $y$-axis represents the true label of the image. BadNet is a small, fully connected neural network.}
\label{fig:confusion_matrices}
\end{figure}

Confusion matrices give an overview of how well a network predicts each class. A confusion matrix for each of the tested models can be seen in \autoref{fig:confusion_matrices}. For both SueezeNet and WoodNet, \textit{Kjartan} is most easily recognized, i.e. least likely to be misclassified. In fact this happened only 19 and 13 times for each network respectively. Furthermore, one can note that for both networks, \textit{Lars} and \textit{Kjartan} is most often misinterpreted to be \textit{Morgan}, with an average of 13 times for the two classes. \textit{Morgan} is most often wrongly predicted to be \textit{Other} by SqueezeNet and \textit{Lars} by WoodNet. Lastly, a small fully connected neural network named BadNet is included to show how a simpler network performs on the data. Note how \textit{Morgan} is its most likely predicted class for all classes except for \textit{Other}.

\begin{figure}
\centering
\includegraphics[width=0.5\textwidth]{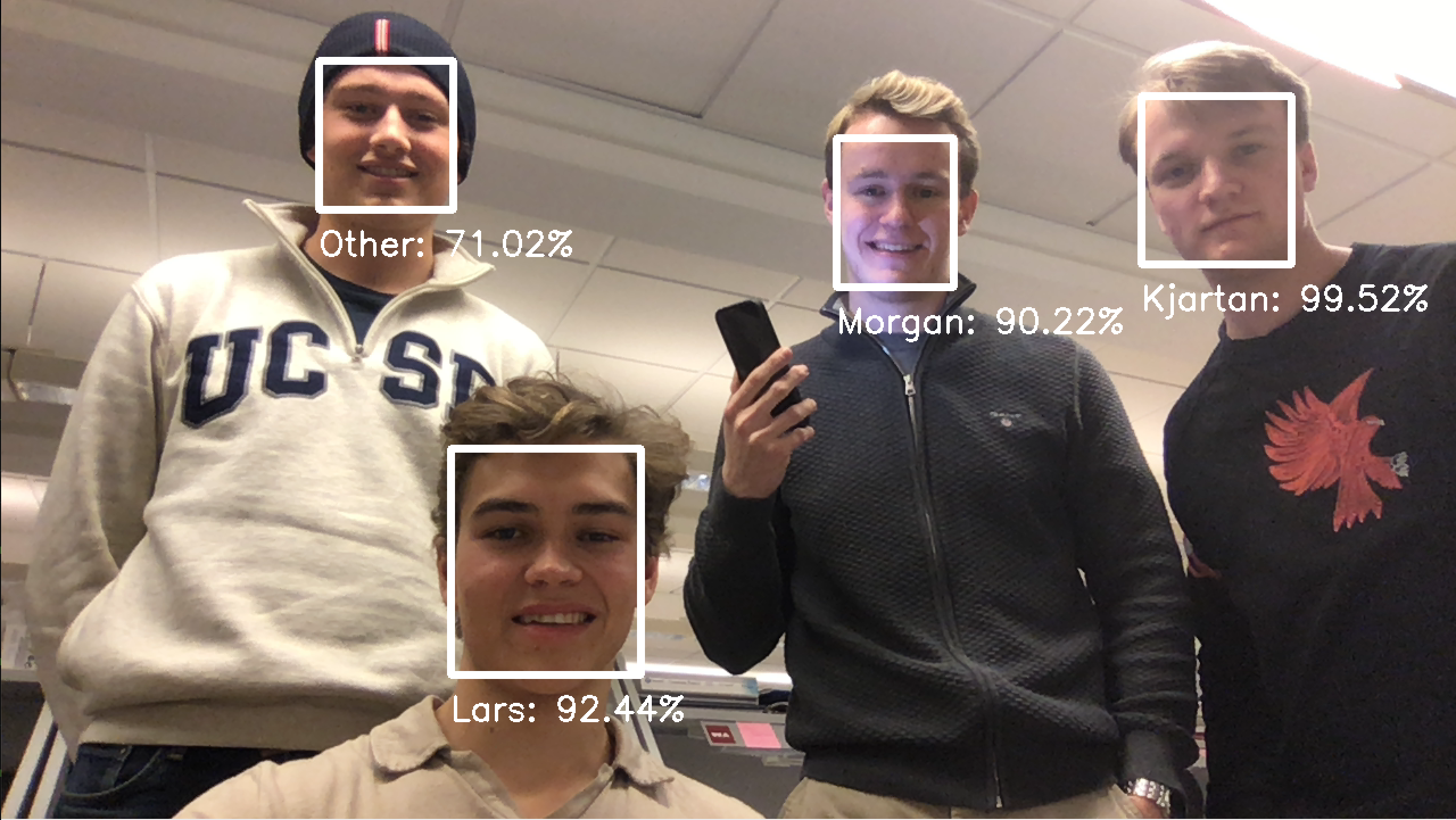}
\caption[Sample screenshot from a live demonstration.]{Sample screenshot of live inference using the SqueezeNet architecture with face detection. Each bounding box represents a detected face, with corresponding predicted label and certainty.}
\label{fig:demo}
\end{figure}

Finally, a live demonstration was developed, running inference in real time using a web camera feed as input. See \autoref{fig:demo} for a sample screenshot. A video including the demonstration is accompanied with this paper and can be referred to for further examination of the performance of the network in a deployed scenario. Qualitatively, the authors were recognized well, however inaccuracies often occurred.

\subsection{Interpretations and Discussion}
As mentioned, \autoref{fig:training-plot} clearly shows that SqueezeNet learns much faster than WoodNet. This seems logical since SqueezeNet is already trained to achieve high accuracy on ImageNet, which is arguably a harder classification task. In theory, that is because SqueezeNet already knows how to effectively extract features from images. By training the last layer (incidentally called ``classifier''), it is only taught how to decide upon a class based on the features it has extracted. In the case of WoodNet, in addition to learning how do this classification, it must also learn how to extract features. Naturally this requires more exposure to images and more resources. Therefore, it seems reasonable to conclude that, as long as it is feasible, one should start with a trained model and adapt it instead of training from scratch.

The confusion matrices in \autoref{fig:confusion_matrices} contain interesting findings. Based on the observations made in \autoref{sec:presentation-of-results}, it seems the two networks do not contain the symmetric property of relations, $R(a,b)$ where $a$ and $b$ are classes and $R$ is ``most similar to''.  For SqueezeNet, this is evident since \textit{Lars} is most often mistaken for \textit{Morgan} and \textit{Morgan} is most regularly misinterpreted as \textit{Other}. As such, we have $R(Lars, Morgan) \centernot\iff  R(Morgan, Lars)$. Therefore, the way SqueezeNet recognizes a face seems to work differently than what humans are used to, assuming that a ``most similar to'' relation is symmetric for humans.

Despite zealous effort, over-fitting seems to be riddling both models. Accuracies on the training, validation, and test set, as seen in \autoref{tab:training-table}, \autoref{tab:key-metrics}, and \autoref{fig:training-plot}, are all very close to 100\%. Even though the authors naturally are happy with the produced model, it must be conceded that this result is most likely too good to be true. Inaccuracies during the live demonstration further indicate this. In the authors' best guess, the cause of over-fit is lack of data, both in terms of variety and numbers. Despite a large number of images, the method of producing these images might be conducive to over-fitting in two ways. The first is that images were extracted from a video recording 30 frames per second, producing similar frames despite constant movement. Furthermore, even though augmentations create different images, they are in many ways still just different views of the same underlying information, and might not be a perfect substitute to actual, unique images. However, as made evident by the video demonstration, the model really does seem to have learned some deeper features of the authors' faces. It is in no way perfect, but it is clear that it performs significantly better than chance, and the authors are more often than not correctly identified.
\section{Conclusion} 
\label{sec:conclusion}

This paper has presented attempted solutions to the problem of facial recognition using convolutional neural networks as a discriminator in entry access control systems. Two proposed solutions in the form of neural network architectures were proposed and compared: building a neural network from scratch and training it using only self-collected data (WoodNet), and leveraging transfer learning by utilizing a network pre-trained for feature extraction of faces and adapting this to our self-collected data (SqueezeNet). This resulted in two models very capable of recognizing the faces of the authors, albeit at the cost of signs of overfitting to the specific data. Both models achieve high accuracy, approaching 100\%, however WoodNet reaches such a level considerably slower during training compared to SqueezeNet as a result of having to learn from scratch.

Considering time constraints and the ambitions of the project, the results seem satisfactory. In particular, the pipeline of data collection, processing, and training the models worked well. However, the quality of the outcome could have been improved through a number of actions. Firstly, as the models seem to be somewhat overfitted to the data, a larger, more diverse dataset could increase generalizability. The methodology for collecting data could have been more rigorous by ensuring less similar frames and using more different settings regarding lighting, background, colors etc. Additionally, collecting more data is always beneficial.

Results kept in mind, deploying the proposed solutions in a entry access control system could present some challenges. It is not entirely unlikely that the models classify strangers as one of the authors, effectively granting unauthorized access. Furthermore, there is little stopping an intruder from displaying an image of one of the authors to gain access. Thus, for such a system to work, further research into improving the models and implementing the access control system may be necessary. One solution could be to combine a camera-based system with other biometric approaches such as fingerprint scanning or voice recognition, or require several successful classifications before granting entry.

Given access to more resources like data, compute, and time, the authors would investigate the possibility of improving the models through use of a larger, more diverse dataset and a deeper network that may be able to extract more detailed features. This would both counteract overfitting as well as improve accuracy of inference. In addition, the evaluation method of \emph{cross-validation} would be leveraged, as it utilizes the data better and provides a valuable metric for the performance of the models, but requires considerable resources. Looking into alternative loss functions that penalize false positives harder than false negatives could also be an effective way to make the system more resilient to intruders.

\newpage
\bibliography{main}

\addappendix
\subsection{Definition of WoodNet CNN in PyTorch Code}
\label{app:woodnet}

\begin{code}
\captionof{listing}{Code that defines the neural network used}
\label{code:woodnet}
\begin{minted}{Python}
class WoodNet(nn.Module):
    size_after_conv = 7 * 7 * 64
    def __init__(self):
        super(CNN, self).__init__()
        self.features = nn.Sequential(   
            nn.Conv2d(3, 32, kernel_size=3, padding=1),
            nn.MaxPool2d(2),
            nn.ReLU(),

            nn.Conv2d(32, 64, kernel_size=3, padding=1),
            nn.MaxPool2d(2),
            nn.ReLU(),
            
            nn.Conv2d(64, 64, kernel_size=3, padding=1),
            nn.MaxPool2d(2),
            nn.ReLU(),
            
            nn.Conv2d(64, 64, kernel_size=3, padding=1),
            nn.MaxPool2d(2),
            nn.ReLU(),
            
            nn.Conv2d(64, 64, kernel_size=3, padding=1),
            nn.MaxPool2d(2),
            nn.ReLU(),
        )
        self.classify = nn.Sequential(
            nn.Linear(self.size_after_conv, 2048),
            nn.ReLU(),
            nn.Linear(2048, 1024),
            nn.ReLU(),
            nn.Dropout(),
            nn.Linear(1024, len(class_names)),
        )

    def forward(self, x):
        x = self.features(x)
        x = x.view(-1, self.size_after_conv)
        x = self.classify(x)

        return x
\end{minted}
\end{code}

\newpage

\subsection{Definition of BadNet in PyTorch Code}
\label{app:badnet}

\begin{code}
\captionof{listing}{Code that defines BadNet, a simple fully connected network used for comparison}
\label{code:badnet}
\begin{minted}{Python}
class BadNet(nn.Module):
    size_after_conv = 7 * 7 * 64
    def __init__(self):
        super(BadNet, self).__init__()
        
        self.classify = nn.Sequential(
            nn.Linear(224*224*3, 128),
            nn.ReLU(),
            nn.Dropout(),
            nn.Linear(128, len(class_names)),
        )

    def forward(self, x):
        x = x.view(-1, 224*224*3)
        x = self.classify(x)

        return x
\end{minted}
\end{code}

\newpage
\subsection{Example Output from the Training Loop after Three Epochs}
\label{app:loop-output}

\begin{center}
  \lstset{%
    caption=Example of the output produced by the training loop,
    basicstyle=\ttfamily\footnotesize\bfseries,
    frame=tb,
    label=lst:loop-output
  }
\begin{lstlisting}
Training only last layer of SqueezeNet
Epoch 0/24
----------
Epoch: 0 (train): 100%|###########| 6836/6836 [04:30<00:00, 25.30it/s]
train Loss: 0.1488 Acc: 0.9476
Epoch: 0 (val): 100%|###########| 1465/1465 [00:51<00:00, 28.62it/s]
val Loss: 0.0499 Acc: 0.9851

Epoch 1/24
----------
Epoch: 1 (train): 100%|###########| 6836/6836 [04:31<00:00, 25.22it/s]
train Loss: 0.0826 Acc: 0.9714
Epoch: 1 (val): 100%|###########| 1465/1465 [00:52<00:00, 28.11it/s]
val Loss: 0.0353 Acc: 0.9900

Epoch 2/24
----------
Epoch: 2 (train): 100%|###########| 6836/6836 [04:43<00:00, 24.14it/s]
train Loss: 0.0704 Acc: 0.9760
Epoch: 2 (val): 100%|###########| 1465/1465 [00:55<00:00, 26.20it/s]
val Loss: 0.0301 Acc: 0.9910

Epoch 3/24
----------
Epoch: 3 (train): 100%|###########| 6836/6836 [04:58<00:00, 22.88it/s]
train Loss: 0.0666 Acc: 0.9766
Epoch: 3 (val):  10%|##         | 141/1465 [00:05<00:50, 26.45it/s]

\end{lstlisting}

\end{center}

\newpage
\subsection{Loss and Accuracy for WoodNet and Squeeznet for Four Epochs}

\begin{table}[H]
\centering
\begin{tabular}{c|l|ll|ll}
\multicolumn{1}{l|}{} &  & \multicolumn{2}{c|}{\textbf{WoodNet}} & \multicolumn{2}{c}{\textbf{SqueezeNet}} \\
\multicolumn{1}{l|}{\textbf{Epoch}} & \textbf{Phase} & \multicolumn{1}{c}{\textbf{Loss}} & \multicolumn{1}{c|}{\textbf{Accuracy}} & \multicolumn{1}{c}{\textbf{Loss}} & \multicolumn{1}{c}{\textbf{Accuracy}} \\ \hline
\multirow{2}{*}{\textbf{1}} & Train & 1.1977 & 0.4230 & 0.1737 & 0.9446 \\
 & Val & 0.5233 & 0.7943 & 0.0709 & 0.9805 \\ \hline
\multirow{2}{*}{\textbf{2}} & Train & 0.2519 & 0.9061 & 0.0863 & 0.9738 \\
 & Val & 0.0815 & 0.9739 & 0.0521 & 0.9849 \\ \hline
\multirow{2}{*}{\textbf{3}} & Train & 0.0626 & 0.9786 & 0.0745 & 0.9765 \\
 & Val & 0.0340 & 0.9884 & 0.0424 & 0.9887 \\ \hline
\multirow{2}{*}{\textbf{4}} & Train & 0.0286 & 0.9900 & 0.0656 & 0.9793 \\ 
 & Val & 0.0249 & 0.9920 & 0.0396 & 0.9894
\end{tabular}
\caption[Loss and accuracy for WoodNet and SqueezeNet for the first four epochs.]{Loss and accuracy for WoodNet and SqueezeNet for the first four epochs of training and validation on the face-cropped dataset.}
\label{tab:training-table}
\end{table}

\end{document}